\documentclass[]{fairmeta}
\usepackage{amsthm,amsmath,amssymb}
\usepackage{graphicx}
\usepackage{natbib}
\usepackage{subcaption}
\usepackage{algorithm,algorithmicx,algpseudocode}
\usepackage{multirow}
\usepackage{booktabs} 
\usepackage[table]{xcolor}

\newtheorem{theorem}{Theorem}[]

\newtheorem{remark1}[theorem]{Remark}

\title{TeleStyle V2: Beyond Content-Preserving Style Transfer with Self-Distillation and Distribution-Matching-Distillation}
\author{Shiwen Zhang}
\author{Yifan Xu}
\author{Haibin Huang}
\author{Chi Zhang}
\author{Xuelong Li}
\affiliation{TeleAI}
\project {\url{https://witcherofresearch.github.io/TeleStyleV2}}
\code{\url{https://github.com/Tele-AI/TeleStyleV2}}
\begin{document}

\abstract{
Given a content reference and a style reference, content-preserving style transfer requires the model to generate stylized outputs with content and  style consistency. We introduced TeleStyle V1 to tackle this problem. However, TeleStyle V1 is trained with photorealistic content reference and artistic style reference, which makes it incapable to cope with artistic content reference and realistic style reference in most cases. In this paper, we designed a Self-Distillation data synthesis strategy to construct such triplets from TeleStyle V1.  Trained with such self-distilled triplets, our TeleStyle V2 supports Content-Style references in the forms of Realistic-and-Realistic (RnR), Realistic-and-Stylized (RnS), Stylized-and-Realistic (SnR), Stylized-and-Stylized (SnS). In addition, we found Distribution Matching Distillation could  preserve the general text-guided image editing capability of the foundation model and fix the content consistency degradation caused by SFT process. Through quantitative evaluations, our TeleStyleV2-QIE-2509-DMD performs at least on par with  Qwen-Image-Edit-2509-DMD, demonstrating strong general image editing skills beyond content-preserving style transfer. We observed the content/style reference order confusion problem in TeleStyle V1 and further introduced prompt enhancer to solve it. TeleStyle V2 uses Qwen-Image-Edit's VLM encoder, Qwen2.5-VL-7B, to generate content prompt and style prompt for free. TeleStyle V2 could achieve comparable style transfer performance with state-of-the-art commercial model, gemini-3-pro-image-preview.

}

\maketitle

\section{Introduction}

In this technical report, we present TeleStyle V2 beyond our previous version, aiming to tackle several issues we observed in TeleStyle V1 \citep{telestyle} and further push the frontier of open-source style transfer models. We observed three major problems in the TeleStyle V1:
\begin{itemize}
    \item {\it The model fails on  stylized content references since  TeleStyle v1 was trained with photo-realistic content references.}
    \item {\it The model fails on realistic style references since TeleStyle v1 was trained with artistic style references in most cases.}
    \item {\it The model confuses the order of content reference and style reference sometimes, returning an image almost identical with style reference.}
\end{itemize}

We introduce a self-distillation mechanism to construct Content-Style-Target triplets.  The Content and Style References could be classified into 4 categories,  Realistic-and-Realistic (RnR), Realistic-and-Stylized (RnS), Stylized-and-Realistic (SnR), Stylized-and-Stylized (SnS). Since such triplets are built with TeleStyle V1, this data synthesis strategy could be considered a self-distillation process. 

\begin{figure}[!ht]
    \centering
     \includegraphics[width=0.9\linewidth]{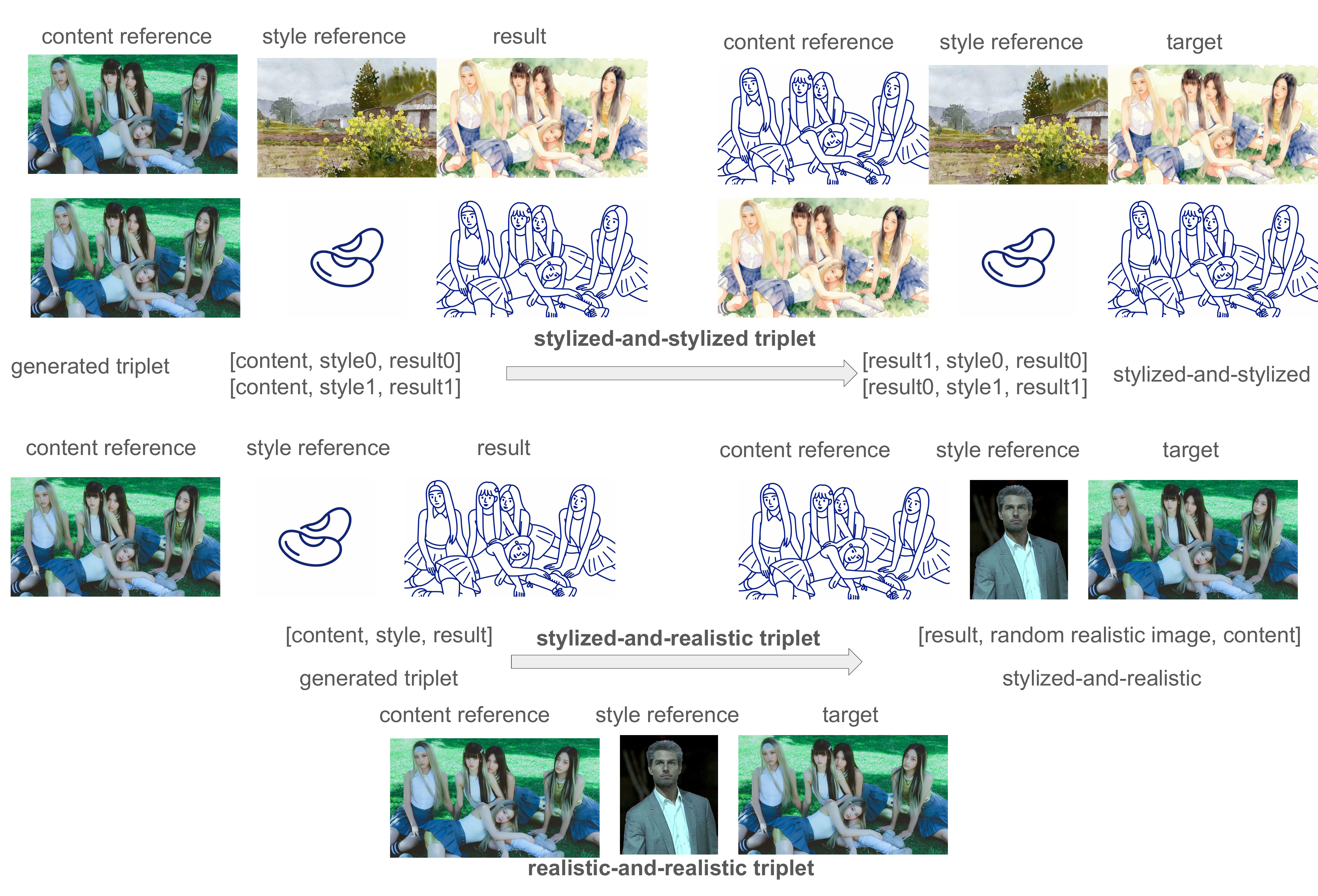}
    \captionsetup{type=figure}
    \captionof{figure}{ We construct self-distillation triplets with TeleStyle v1,  to enable TeleStyle v2 coping with stylized content reference and realistic style reference, beyond realistic content reference and stylized style reference setting in TeleStyle v1. Please note that due to privacy concerns, we use idols and actors' pictures in this figure  for demonstration. They {\bf don't} exist in our training set. Our training set is built with amateurs photos, which are not shown due to privacy reasons.}
    \label{figure_selfdistilltriplet}
    
\end{figure}
In TeleStyle v1 and v2, we found Distribution Matching Distillation (DMD) \citep{dmd2,phasedmd} critical for preserving the content characteristics and helpful to avoid catastrophic forgetting. With one stone two birds, DMD prevents characteristics shifting and decreases inference steps. DMD enables TeleStyle V1 and V2 to perform as well as (if not better than) vanilla Qwen-Image-Edit series (2509,2511) \citep{wu2025qwen} on GEdit-EN and GEdit-CN \citep{stepfunedit}. Such a generalization capability indicates our TeleStyle could be used as a general text-guided image editing  model \citep{imagic,zhang2023forgedit,fastimagic,wu2025qwen,gpt4o,gemini3pro}, beyond style transfer task.  

In addition, we observed a phenomenon we termed {\it Reference Order Confusion}, where for some cases, TeleStyle generates images almost identical with style reference, showing no relation with the content reference, despite MSRoPE \citep{rope,wu2025qwen} applied in Qwen-Image-Edit and image indices used in input template. We tried Direct Preference Optimization (DPO) \citep{dpo} to tackle this problem, which, surprisingly, did not work. We solved this problem via prompt enhancing using  VLM \citep{qwen2.5vl}.

\vspace{5mm}
Our main contributions are:

\begin{itemize}
   \item We introduced a Self-Distillation mechanism to construct Content-and-Style References of Realistic-and-Realistic (RnR), Realistic-and-Stylized (RnS), Stylized-and-Realistic (SnR), Stylized-and-Stylized (SnS). Trained with such triplets, TeleStyle V2 is strengthened in its generalization capability.
   \item We found utilizing Distribution-Matching-Distillation (DMD) mitigates the content consistency degradation problem and further makes TeleStyle  a general text-guided image editor on par with Qwen-Image-Editing series. 
   \item We observed the Reference Order Confusion problem and tackle it with prompt enhancement by adding  a brief content description to the prompt.
   
\end{itemize}

For the content-preserving style transfer task, TeleStyle V2 achieves style similarity and content consistency on par with the state-of-the-art commercial model gemini-3-pro-image-preview, i.e. nano banana pro \citep{gemini3pro}.

\section{Methods}

\subsection{Connecting Different Style Domains via Self-Distillation}
\label{sectiondata}

\subsubsection{Brief Recap of Training Triplets in TeleStyle V1}
In TeleStyle V1, we introduced how we constructed the triplet dataset [content ref, style ref, target]. The dataset is mixed with collected triplets $D_{collect}$ \citep{gpt4o,song2025omniconsistency} and synthetic triplets $D_{synthetic}$\citep{li2024styletokenizer,zhang2025cdst}. We proposed Style-CCL \citep{styleccl,qwenstyle}, a curriculum continual learning framework \citep{bengio2009curriculum,zhang2020knowledge,zhang2022tfcnet,zhangv4d} to enable Qwen-Image-Edit-2509 \citep{wu2025qwen} performing content-preserving style transfer task. However, in the training triplets, the content references are all realistic photographs, most of the style references are artistic images, i.e. the content reference and style reference are {\bf R}ealistic-and-{\bf S}tylized (RnS). Thus TeleStyle v1 is incapable of changing the style of a stylized content reference in most cases.  
\subsubsection{Direct Connection for Realism and Stylization}
We categorize an image into binary classes, either realistic or stylized. Thus we could classify the content and style references into 4 categories:  Realistic-and-Realistic (RnR), Realistic-and-Stylized (RnS), Stylized-and-Realistic (SnR), Stylized-and-Stylized (SnS). In TeleStyle v1, we only considered the RnS setting during dataset construction.  In Figure \ref{figure_selfdistilltriplet}, we show that it is quite easy to construct RnR triplets by using the same realistic image as content reference and target, then assigning a random realistic image as style reference. However, for SnS and SnR triplets, it is not very straight-forward. Since TeleStyle v1 is already a state-of-the-art content-preserving style transfer model, we introduce a self-distillation mechanism to construct SnS and SnR triplets. 

For SnS, demonstrated in Figure \ref{figure_selfdistilltriplet}, given the same content reference, we use two random style references to generate two targets, with the order of {\it[content ref, style ref, target]}:

\begin{equation}
[content,style_0,result_0]\longrightarrow [result_1,style_0,result_0], 
\end{equation}
\begin{equation}
    [content,style_1,result_1]\longrightarrow[result_0,style_1,result_1].
\end{equation}
Thus, we obtain two triplets, where the content reference,  style references, and the targets, are all stylized images.

For SnR, in Figure \ref{figure_selfdistilltriplet}, we have
\begin{equation}
    [content,style,result]\longrightarrow[result, a\ random\ realistic\ image,content].
\end{equation}

The construction of training triplets utilized the existing capability itself, thus this is a self-distillation approach. Through further experiments, we found the potential of further data mining, which we leave for future exploration. Via self-distillation, we bridge  and connect the disjoint realistic and stylized modalities.  We merge these new triplets into our training set and train TeleStyle v2 with Style-CCL. However, we found that the SnR task is very challenging for reference-guided generation and the performance is very instable. Thus we recommend directly using the prompt "convert Figure 1 to a photorealistic photograph" on the content reference instead of using the content-preserving style transfer mode.

\subsection{Distribution Matching Distillation is the Savior}

\begin{figure*}[t]
  \centering
  \includegraphics[width=1\linewidth]{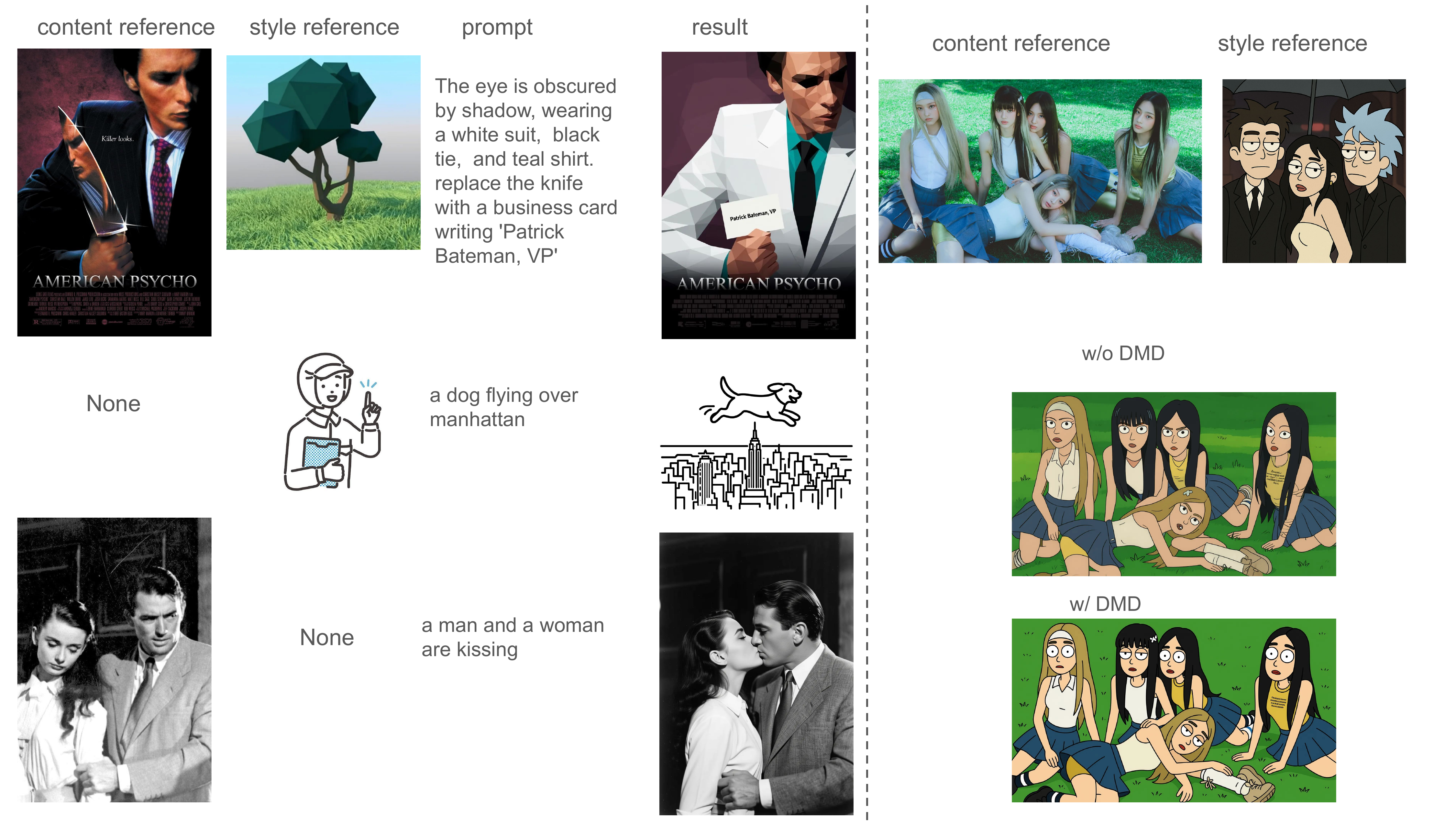}
  \caption{DMD helps TeleStyle to preserve the general image editing capability of Qwen-Image-Edit series and mitigate the degradation caused by style transfer SFT.}
  \vspace{-0.2cm}
  \label{figure_dmd}
\end{figure*}

The training of TeleStyle leads to certain content consistency degradation and image distortion to the  Qwen-Image-Edit foundation models, shown in Figure \ref{figure_dmd}. Though this is far from "catastrophic" forgetting, we could still fix such degradation via minimizing the reverse Kullback-Leibler (KL) Divergence $D_{KL}(p_{fake} \| p_{real})$ \citep{kld}. With the real data distribution ${\displaystyle p_{real}(x_0)}$ and the generated data distribution ${\displaystyle p_{fake}(x_0)}$ produced by ${\boldsymbol{G}_{\boldsymbol{\phi}}}$, the gradient of reverse KL Divergence is:
\begin{equation}
% \begin{aligned}
\nabla_{\boldsymbol{\phi}} D_{KL}(p_{fake} \| p_{real}) = E_{z, x_0 = \boldsymbol{G}_{\boldsymbol{\phi}}(z)} [(\nabla_{x_0}\log p_{fake}(x_0) - \nabla_{x_0} \log p_{real}(x_0)) \frac{d \boldsymbol{G}}{d \boldsymbol{\phi}} ] 
\label{eq:reverse_KL}
% \end{aligned}
\end{equation} 
where $z \sim \mathcal{N}(0, \boldsymbol{I})$ is a random Gaussian noise input. 

With Distribution Matching Distillation (DMD) \citep{dmd2,phasedmd,lightx2v} models on Qwen-Image-Series applied to TeleStyle, we could hit two birds with one stone. The inference cost is reduced 10 times, and the content consistency and vanilla image editing capability of Qwen-Image-Editing series are preserved, even strengthened.   TeleStyle could conduct content-preserving style transfer and text-guided image editing in one run ( the first example in Figure \ref{figure_dmd}). In addition, TeleStylev2 accepts optional number of reference images. For example, it works on style reference + prompt scenario by providing only one style reference. It also works on traditional text-guided image editing task by providing one content reference. We also compare TeleStyle v2 with and without DMD in Figure \ref{figure_dmd}. TeleStyle v2 works without DMD, though the aesthetics could drop and the image could be distorted sometimes. It is worth noting that TeleStyle v2 boosts Qwen-Image-Edit series on style-related tasks for text-guided image editing, even without being trained on this text-guided task.

\subsection{Tackling Reference Order Confusion}
We observed TeleStyle V1 sometimes confuses the order of content and style reference in a certain way so that the generated result is very similar to the style reference and has no relation with content reference. We term such a phenomenon {\it Reference Order Confusion}. Though Qwen-Image-Edit series have already applied MS-RoPE to distinguish the references and further encoded image indices in the VLM input template, the problem persists for some cases. To tackle the Reference Order Confusion,   We tried Direct Preference Optimization (DPO) \citep{dpo}, and surprisingly found it not working. We finally solve this problem by adding auto-generated content description to the prompt. Such a simple hint inspires the model to understand the actual target.
\subsubsection{Direct Preference Optimization Does Not Work}
We first tried Direct Preference Optimization (DPO) to mitigate  the reference order confusion. We tried two data scales of  win-lose training pairs [$x^{win}_0,x^{lose}_0$]: 2k and 400k. The 2k data is obtained by manually filtering win-lose pairs from a 100k fraction of 10 million generated results of TeleStyle V1. The 400k data is constructed from triplet data of section \ref{sectiondata}, where we use target image as the winning image and style reference as the lost image. The loss of DPO is:

\begin{equation}
\label{equ:fm_dpo_full}
\left\{
\begin{aligned}
& \text{Diff}_{\text{policy}} = 
     \left( \big\| v_\theta(x_t^{win}, h, t) - v_t^{win} \big\|_2^2 - \big\| v_\theta(x_t^{lose}, h, t) - v_t^{lose} \big\|_2^2 \right) \\
& \text{Diff}_{\text{ref}} = 
     \left( \big\| v_{\text{ref}}(x_t^{win}, h, t) - v_t^{win} \big\|_2^2 - \big\| v_{\text{ref}}(x_t^{lose}, h, t) - v_t^{lose} \big\|_2^2 \right) \\
& \mathcal{L}_{DPO} = -\mathbb{E}_{h, (x_0^{win}, x_0^{lose}) \sim \mathcal{D},\, t \sim \mathcal{U}(0, 1)}
\Bigg[
      \log \sigma\Big( -\beta\, (\text{Diff}_{\text{policy}} - \text{Diff}_{\text{ref}}) \Big)
\Bigg],
\end{aligned}
\right.
\end{equation}
where $\text{Diff}_{\text{policy}}$ and $\text{Diff}_{\text{ref}}$ denote the preference differences computed by the policy model $v_\theta$ and the reference model $v_{ref}$, respectively, $\beta$ is a scaling parameter, and $\sigma(\cdot)$ denotes the sigmoid function. $h$ denotes the vision-text embedding from VLM,  $x_0^{win}$ denotes the VAE \citep{vae} latent of  winning image, and $x_0^{lose}$ for the losing image. Timestep is sampled from $t\sim (0, 1)$ to construct the input latent variable $x_t^{win}$ and $x_t^{lose}$ as well as their corresponding velocity $v_t^{win}$ and $v_t^{lose}$.

We trained a DPO Lora on Qwen-Image-Edit-2509. Unfortunately and quite surprisingly, DPO could not solve this reference order confusion problem, not even with 400K data. 
\subsubsection{Prompt Enhancement: The Solution}
 With further investigating, we found a very simple solution: with a very brief description of the objects in the content reference with a few separated words, the model immediately understands the editing purpose. Thus we use the Qwen2.5-VL-7B \citep{qwen2.5vl} to complete this task, which has been already loaded into the memory as the visual-language encoder of Qwen-Image-Edit model. The generated words are called content prompt. We further use Qwen2.5-VL-7B to describe the style reference, which is called style prompt. The content and style prompts are optional during inference, with the content prompt   solving the Reference Order Confusion problem and style prompt might solving some corner cases to enhance style similarity.

\begin{figure*}[!htb]
  \centering
  %\hspace{-10mm}
  %\vspace{-0.7cm}
   \includegraphics[width=0.9\linewidth]{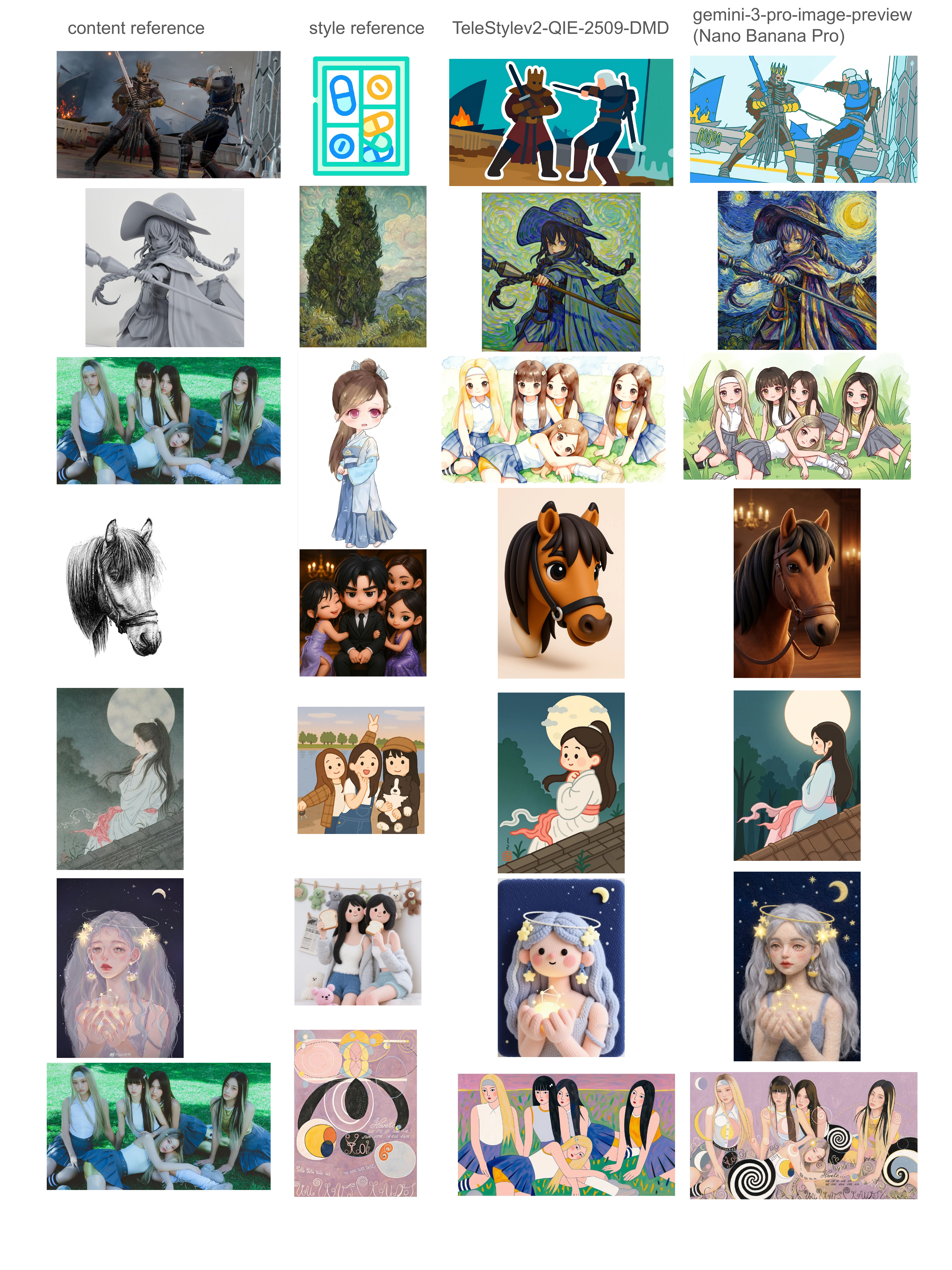}%{egfigure.eps}
   %\vspace{-0.4cm}
   \caption{ Qualitative Comparison of TeleStylev2-QIE-2509-DMD with the state-of-the-art image generation model, gemini-3-pro-image-preview (nano-banana-pro).   }
   \label{figure_compare}
\end{figure*}
\section{Experiments and Evaluation}

\subsection{Implementation}
%image starts here
 We trained TeleStyle v2 on Qwen-Image-Edit-2509, Qwen-Image-Edit-2511 \citep{wu2025qwen} and FLUX 2 Klein \citep{flux-2-2025}, which are foundation models based  on mmdit structure \citep{dit,sd3,flux2024,flux-2-2025} instead of UNet \citep{ddpm,podell2023sdxl}. We trained both Lora \citep{lora} and full-parameter models \citep{wu2025qwen} on these foundation models. However, all the qualitative and quantitative results shown in this paper are produced by TeleStyleV2-QIE-2509 Lora. The optimizer is AdamW \citep{adam,adamw}, and the learning rate is 1e-4. The Loras are trained on 4 H100 with gradient checkpointing \citep{griewank2000algorithm} and full-parameter models on 8 H100 with DeepSpeed \citep{rasley2020deepspeed}.

 \subsection{Evaluation on General Image Editing Capability}
 We test the general image editing capability of TeleStyle v2 on GEdit-Bench \citep{stepfunedit} evaluated by Qwen2.5-VL-72B \citep{qwen2.5vl}. Shown in Table \ref{table_gedit}, with Qwen-Image-Edit-2509 (QIE-2509) as the base model,  TeleStylev2-DMD demonstrates comparable text-guided image editing performance with QIE-2509-DMD on 11 different sub-tasks. Such metrics prove that beyond  strong content-preserving style transfer capability, TeleStylev2-DMD  also demonstrates competitive text-guided  image editing performance on par with vanilla Qwen-Image-Edit. Though our training triplets are all English,  TeleStylev2 could even outperform vanilla QIE-2509-DMD on GEdit-Chinese. Specifically, there is a noticeable improvement on the style-changing sub-task, though we did not train TeleStylev2 on any text-guided style transfer data.

\begin{table*}[ht]
\centering
\resizebox{\textwidth}{!}{%
\begin{tabular}{l|ccc|ccc|ccc|ccc}
\toprule
\multirow{3}{*}{\textbf{Model}} 
& \multicolumn{6}{c|}{\textbf{GEdit-EN (Full set)} $\uparrow$} 
& \multicolumn{6}{c}{\textbf{GEdit-CN (Full set)} $\uparrow$} \\
\cmidrule(lr){2-7} \cmidrule(lr){8-13}
& \multicolumn{3}{c|}{QIE-2509-DMD}  & \multicolumn{3}{c|}{TeleStylev2-QIE-2509-DMD} & \multicolumn{3}{c|}{QIE-2509-DMD}  & \multicolumn{3}{c}{TeleStylev2-QIE-2509-DMD}\\
\cmidrule(lr){2-7} \cmidrule(lr){8-13}
& \textbf{Q\_SC} & \textbf{Q\_PQ} & \textbf{Q\_O} 
& \textbf{Q\_SC} & \textbf{Q\_PQ} & \textbf{Q\_O} 
& \textbf{Q\_SC} & \textbf{Q\_PQ} & \textbf{Q\_O} 
& \textbf{Q\_SC} & \textbf{Q\_PQ} & \textbf{Q\_O} \\
\midrule
background change
&   8.050& 7.700& 7.866 & 8.150& 7.650& 7.836 & 8.000& 7.700& 7.838 &8.050& 7.675& 7.800 \\
color alter
&  8.500& 7.300& 7.843 & 8.225& 7.250& 7.570 &8.275& 7.225& 7.683 &  8.275& 7.275& 7.709\\
material alter
& 6.850& 7.025& 6.386 &  7.075& 6.975& 6.538 &6.800& 6.725& 6.259 &6.825& 7.025& 6.347 \\
motion change &7.425& 7.750& 7.467 &7.550& 7.800& 7.545 & 7.475& 7.650& 7.420 & 7.325& 7.775& 7.326\\
ps human &6.714& 7.529& 6.711 &6.571& 7.543& 6.539 &6.500& 7.543& 6.555 & 6.543& 7.600& 6.618\\
\rowcolor{teal!20} style change &6.583& 7.183& 6.479 &7.133& 7.083& {\bf 6.943} & 6.867& 7.167& 6.606 & 7.267& 7.083& {\bf 7.053}\\
subject add &7.950& 7.667& 7.720 & 7.967& 7.617& 7.697 & 8.050& 7.583& 7.761 & 7.983& 7.717& 7.832\\
subject remove &8.263& 7.667& 7.803 & 8.123& 7.719& 7.752 & 7.561& 7.667& 7.185 &8.158& 7.754& 7.794\\
subject replace &8.317& 7.483& 7.865 &  8.217& 7.567& 7.803 &8.083& 7.483& 7.691 & 8.150& 7.533& 7.816\\
text change &9.051& 7.525& 8.225 & 9.020& 7.505& 8.157 &8.747& 7.485& 8.013 & 8.859& 7.505& 8.121\\
tone transfer &7.150& 7.400& 7.136 &7.075& 7.500& 6.978 & 7.075& 7.325& 6.886 & 7.400& 7.350& 7.284\\
\midrule
\textbf{avg} 
& 7.714& 7.475& 7.409&7.737& 7.474& 7.396&7.585& 7.414& 7.263&7.712& 7.481& 7.427
 \\
\bottomrule
\end{tabular}%
}
\vspace{10pt}
\caption{\textbf{Quantitative evaluation on GEdit-EN and GEdit-CN.} We evaluate Qwen-Image-Edit-2509-DMD (QIE-2509-DMD) and TeleStylev2-QIE-2509-DMD from all aspects of GEdit-English and GEdit-Chinese instructions. TeleStylev2 performs on par with, or even better than QIE-2509-DMD on such general text-guided image editing tasks, with specific improvement on the style-related task.   Semantic Consistency (Q\_SC), Perceptual Quality (Q\_PQ), and Overall Score (Q\_O) refer to the metrics evaluated by Qwen2.5-VL-72B.
}
\label{table_gedit}
\vspace{-1pt}
\end{table*}

\subsection{Comparison with gemini-3-pro-image-preview}

We compare TeleStylev2-QIE-2509-DMD with the state-of-the-art image generation model gemini-3-pro-image-preview (nano banana pro) \citep{gemini3pro} in Figure \ref{figure_compare}. Our model outperforms gemini-3-pro-image-preview in terms of style similarity in most cases and on par with it in terms of content consistency. We choose not to release the quantitative comparison and user study because we found TeleStyle V2's win rate against nano-banana-pro is very high, which is quite unexpected and we suspect that our testing set might be biased.  To make a safe statement, TeleStyleV2-QIE2509-DMD almost performs on par with gemini-3-pro-image-preview on content-preserving style transfer task.

\section{Conclusion}
In this work, we introduce TeleStyle V2, improving the content-preserving style transfer performance beyond TeleStyle V1 (QwenStyle). TeleStyle V2 introduces self-distillation to improve generalization and employs distribution-matching-distillation to preserve the general image editing capability of foundation models and to repair the content consistency degradation caused by SFT. Prompt Enhancer instead of DPO solves the reference order confusion problem and further solves some corner cases. 

The training of TeleStyle V2 was finished in March, 2026. In the near future, we will release a multi-reference subject-style driven Omnimodel, based on self-distillation of TeleStyle and TeleComposer \citep{telecomposer}.

{
\bibliographystyle{ieee}
\bibliography{paper}

@String(CVPR= {IEEE Conf. Comput. Vis. Pattern Recog.})

@String(ICLR = {Int. Conf. Learn. Represent.})

@String(AAAI = {AAAI})

@String(CVPR  = {CVPR})

@String(ICLR  = {ICLR})

@inproceedings{imagic,
  author       = {Bahjat Kawar and
                  Shiran Zada and
                  Oran Lang and
                  Omer Tov and
                  Huiwen Chang and
                  Tali Dekel and
                  Inbar Mosseri and
                  Michal Irani},
  title        = {Imagic: Text-Based Real Image Editing with Diffusion Models},
  booktitle    = {{CVPR}},
  pages        = {6007--6017},
  publisher    = {{IEEE}},
  year         = {2023}
}

@inproceedings{adam,
  author       = {Diederik P. Kingma and
                  Jimmy Ba},
  title        = {Adam: {A} Method for Stochastic Optimization},
  booktitle    = {{ICLR} (Poster)},
  year         = {2015}
}

@inproceedings{vae,
  author       = {Diederik P. Kingma and
                  Max Welling},
  title        = {Auto-Encoding Variational Bayes},
  booktitle    = {{ICLR}},
  year         = {2014}
}

@article{zhang2023forgedit,
  title={Forgedit: Text guided image editing via learning and forgetting},
  author={Zhang, Shiwen and Xiao, Shuai and Huang, Weilin},
  journal={arXiv preprint arXiv:2309.10556},
  year={2023}
}

@inproceedings{zhangv4d,
  title={V4D: 4D Convolutional Neural Networks for Video-level Representation Learning},
  author={Zhang, Shiwen and Guo, Sheng and Huang, Weilin and Scott, Matthew R and Wang, Limin},
  booktitle={International Conference on Learning Representations},
  year         = {2020}
}

@inproceedings{zhang2020knowledge,
  title={Knowledge integration networks for action recognition},
  author={Zhang, Shiwen and Guo, Sheng and Wang, Limin and Huang, Weilin and Scott, Matthew},
  booktitle={Proceedings of the AAAI Conference on Artificial Intelligence},
  year={2020}
}

@article{zhang2022tfcnet,
  title={Tfcnet: Temporal fully connected networks for static unbiased temporal reasoning},
  author={Zhang, Shiwen},
  journal={arXiv preprint arXiv:2203.05928},
  year={2022}
}

@inproceedings{
fastimagic,
title={{Fast Imagic: Solving Overfitting in Text-guided Image Editing via Disentangled UNet with Forgetting Mechanism and Unified Vision-Language Optimization}},
author={Zhang, Shiwen},
booktitle={PMLR},
year={2024},
}

@article{podell2023sdxl,
  title={Sdxl: Improving latent diffusion models for high-resolution image synthesis},
  author={Podell, Dustin and English, Zion and Lacey, Kyle and Blattmann, Andreas and Dockhorn, Tim and M{\"u}ller, Jonas and Penna, Joe and Rombach, Robin},
  journal={arXiv preprint arXiv:2307.01952},
  year={2023}
}

@misc{flux2024,
    author={Black Forest Labs},
    title={FLUX},
    year={2024},
    howpublished={\url{https://github.com/black-forest-labs/flux}},
}

@inproceedings{dit,
  title={Scalable diffusion models with transformers},
  author={Peebles, William and Xie, Saining},
  booktitle={Proceedings of the IEEE/CVF International Conference on Computer Vision},
  pages={4195--4205},
  year={2023}
}

@article{wu2025qwen,
  title={Qwen-image technical report},
  author={Wu, Chenfei and Li, Jiahao and Zhou, Jingren and Lin, Junyang and Gao, Kaiyuan and Yan, Kun and Yin, Sheng-ming and Bai, Shuai and Xu, Xiao and Chen, Yilei and others},
  journal={arXiv preprint arXiv:2508.02324},
  year={2025}
}

@article{song2025omniconsistency,
  title={Omniconsistency: Learning style-agnostic consistency from paired stylization data},
  author={Song, Yiren and Liu, Cheng and Shou, Mike Zheng},
  journal={arXiv preprint arXiv:2505.18445},
  year={2025}
}

@article{lora,
  title={Lora: Low-rank adaptation of large language models},
  author={Hu, Edward J and Shen, Yelong and Wallis, Phillip and Allen-Zhu, Zeyuan and Li, Yuanzhi and Wang, Shean and Wang, Lu and Chen, Weizhu},
  journal={arXiv preprint arXiv:2106.09685},
  year={2021}
}

@inproceedings{li2024styletokenizer,
  title={Styletokenizer: Defining image style by a single instance for controlling diffusion models},
  author={Li, Wen and Fang, Muyuan and Zou, Cheng and Gong, Biao and Zheng, Ruobing and Wang, Meng and Chen, Jingdong and Yang, Ming},
  booktitle={European Conference on Computer Vision},
  pages={110--126},
  year={2024},
  organization={Springer}
}

@article{rope,
  title={Roformer: Enhanced transformer with rotary position embedding},
  author={Su, Jianlin and Ahmed, Murtadha and Lu, Yu and Pan, Shengfeng and Bo, Wen and Liu, Yunfeng},
  journal={Neurocomputing},
  volume={568},
  pages={127063},
  year={2024},
  publisher={Elsevier}
}

@article{ddpm,
  title={Denoising diffusion probabilistic models},
  author={Ho, Jonathan and Jain, Ajay and Abbeel, Pieter},
  journal={Advances in neural information processing systems},
  volume={33},
  pages={6840--6851},
  year={2020}
}

@inproceedings{bengio2009curriculum,
  title={Curriculum learning},
  author={Bengio, Yoshua and Louradour, J{\'e}r{\^o}me and Collobert, Ronan and Weston, Jason},
  booktitle={Proceedings of the 26th annual international conference on machine learning},
  pages={41--48},
  year={2009}
}

@article{griewank2000algorithm,
  title={Algorithm 799: revolve: an implementation of checkpointing for the reverse or adjoint mode of computational differentiation},
  author={Griewank, Andreas and Walther, Andrea},
  journal={ACM Transactions on Mathematical Software (TOMS)},
  volume={26},
  number={1},
  pages={19--45},
  year={2000},
  publisher={ACM New York, NY, USA}
}

@article{gpt4o,
  title={Gpt-4o system card},
  author={Hurst, Aaron and Lerer, Adam and Goucher, Adam P and Perelman, Adam and Ramesh, Aditya and Clark, Aidan and Ostrow, AJ and Welihinda, Akila and Hayes, Alan and Radford, Alec and others},
  journal={arXiv preprint arXiv:2410.21276},
  year={2024}
}

@article{zhang2025cdst,
  title={CDST: Color Disentangled Style Transfer for Universal Style Reference Customization},
  author={Zhang, Shiwen and Chen, Zhuowei and Chen, Lang and Wu, Yanze},
  journal={arXiv preprint arXiv:2506.13770},
  year={2025}
}

@article{qwen2.5vl,
  title={Qwen2. 5-vl technical report},
  author={Bai, Shuai and Chen, Keqin and Liu, Xuejing and Wang, Jialin and Ge, Wenbin and Song, Sibo and Dang, Kai and Wang, Peng and Wang, Shijie and Tang, Jun and others},
  journal={arXiv preprint arXiv:2502.13923},
  year={2025}
}

@inproceedings{sd3,
  title={Scaling rectified flow transformers for high-resolution image synthesis},
  author={Esser, Patrick and Kulal, Sumith and Blattmann, Andreas and Entezari, Rahim and M{\"u}ller, Jonas and Saini, Harry and Levi, Yam and Lorenz, Dominik and Sauer, Axel and Boesel, Frederic and others},
  booktitle={Forty-first international conference on machine learning},
  year={2024}
}

@article{dmd2,
  title={Improved distribution matching distillation for fast image synthesis},
  author={Yin, Tianwei and Gharbi, Micha{\"e}l and Park, Taesung and Zhang, Richard and Shechtman, Eli and Durand, Fredo and Freeman, William T},
  journal={Advances in neural information processing systems},
  volume={37},
  pages={47455--47487},
  year={2024}
}

@article{gemini3pro,
  title={Gemini-3-Pro-Image-Preview},
  author={Gemini Team},
  year={2025}
}

@inproceedings{phasedmd,
  title={Phased dmd: Few-step distribution matching distillation via score matching within subintervals},
  author={Fan, Xiangyu and Qiu, Zesong and Wu, Zhuguanyu and Wang, Fanzhou and Lin, Zhiqian and Ren, Tianxiang and Lin, Dahua and Gong, Ruihao and Yang, Lei},
  booktitle={Proceedings of the IEEE/CVF Conference on Computer Vision and Pattern Recognition},
  pages={41667--41676},
  year={2026}
}

@article{dpo,
  title={Direct preference optimization: Your language model is secretly a reward model},
  author={Rafailov, Rafael and Sharma, Archit and Mitchell, Eric and Manning, Christopher D and Ermon, Stefano and Finn, Chelsea},
  journal={Advances in neural information processing systems},
  volume={36},
  pages={53728--53741},
  year={2023}
}

@article{telestyle,
  title={TeleStyle: Content-Preserving Style Transfer in Images and Videos},
  author={Zhang, Shiwen and Yang, Xiaoyan and Zi, Bojia and Huang, Haibin and Zhang, Chi and Li, Xuelong},
  journal={arXiv preprint arXiv:2601.20175},
  year={2026}
}

@article{styleccl,
  title={Style-CCL: Content-Preserving Style Transfer  via Curriculum Continual Learning},
  author={Zhang, Shiwen and Wang, Haoyuan and Zang, Xianghao and Huang, Haibin and Zhang, Chi and Li, Xuelong},
  journal={arXiv preprint arXiv:2606},
  year={2026}
}

@article{qwenstyle,
  title={QwenStyle: Content-Preserving Style Transfer with Qwen-Image-Edit},
  author={Zhang, Shiwen and Huang, Haibin and Zhang, Chi and Li, Xuelong},
  journal={arXiv preprint arXiv:2601.06202},
  year={2026}
}

@misc{lightx2v,
author = {LightX2V Contributors},
title = {LightX2V: Light Video Generation Inference Framework},
year = {2025},
publisher = {GitHub},
journal = {GitHub repository},
howpublished = {\url{https://github.com/ModelTC/lightx2v}},
}

@article{stepfunedit,
  title={Step1x-edit: A practical framework for general image editing},
  author={Liu, Shiyu and Han, Yucheng and Xing, Peng and Yin, Fukun and Wang, Rui and Cheng, Wei and Liao, Jiaqi and Wang, Yingming and Fu, Honghao and Han, Chunrui and others},
  journal={arXiv preprint arXiv:2504.17761},
  year={2025}
}

@article{adamw,
  title={Decoupled weight decay regularization},
  author={Loshchilov, Ilya and Hutter, Frank},
  journal={arXiv preprint arXiv:1711.05101},
  year={2017}
}

@inproceedings{rasley2020deepspeed,
  title={Deepspeed: System optimizations enable training deep learning models with over 100 billion parameters},
  author={Rasley, Jeff and Rajbhandari, Samyam and Ruwase, Olatunji and He, Yuxiong},
  booktitle={Proceedings of the 26th ACM SIGKDD international conference on knowledge discovery \& data mining},
  pages={3505--3506},
  year={2020}
}

@article{telecomposer,
  title={TeleComposer: Multi-Reference-Driven Image and Video Customization},
  author={Zhang, Shiwen and Xu, Yifan and Huang, Haibin and Zhang, Chi and Li, Xuelong},
  journal={arXiv preprint arXiv:2606},
  year={2026}
}

@misc{flux-2-2025,
    author={Black Forest Labs},
    title={{FLUX.2: Frontier Visual Intelligence}},
    year={2025},
    howpublished={\url{https://bfl.ai/blog/flux-2}},
}

@article{kld,
  title={On information and sufficiency},
  author={Kullback, Solomon and Leibler, Richard A},
  journal={The annals of mathematical statistics},
  volume={22},
  number={1},
  pages={79--86},
  year={1951},
  publisher={JSTOR}
}
}
\end{document}